\documentclass[runningheads]{llncs}

\usepackage{eccv}

\usepackage{eccvabbrv}

\usepackage{graphicx}
\usepackage{booktabs}
\usepackage{multirow}

\usepackage[accsupp]{axessibility}  

\usepackage{xcolor}
\usepackage{pifont}

\usepackage{tikz}

\definecolor{algogreen}{HTML}{499B4A}
\definecolor{algored}{HTML}{D12A2F}

\usepackage{hyperref}

\usepackage{orcidlink}

\begin{document}

\title{Towards Neuro-Symbolic Procedural Reasoning for Long-Horizon Vision-Language-Action Manipulation} 

\titlerunning{Neuro-Symbolic Procedural Reasoning for Long-Horizon VLA}

\author{
Vivek Chavan\inst{1,2}\orcidlink{0000-0001-9350-5259} \thanks{Project Lead. Correspondence: \texttt{\href{mailto:contact@vivekchavan.com}{contact@vivekchavan.com}}} \and
Yahuan Shi\inst{2} \and
Oliver Heimann\inst{1} \and
Kevin Haninger\inst{1} \and
Jörg Krüger\inst{1,2}\orcidlink{0000-0001-5138-0793} 
}

\authorrunning{V. Chavan et al.}

\institute{
Fraunhofer IPK, Berlin, Germany\\
\and
Technische Universität Berlin, Berlin, Germany
}

\maketitle

\begin{abstract}
Vision-language-action (VLA) models can execute short manipulation skills, but remain brittle in long-horizon procedures requiring persistent task state, dependency-aware reasoning, conditional decisions, and reliable grounding. We investigate a neuro-symbolic framework that combines learned VLA control with explicit task graphs and multimodal procedural memory. Task graphs encode action dependencies, valid transitions, and branch conditions, while memory maintains the active step, completed actions, textual context, and task-relevant visual evidence. Together, these structures guide object selection, destination grounding, subgoal dispatch, and verification of expected state transitions. Human demonstrations provide additional spatial and temporal guidance through gaze or saliency cues. To isolate their effect on policy learning, our initial study bypasses cross-view gaze transfer and directly annotates pseudo-gaze in robot-view teleoperation videos. The resulting guidance is used during VLA fine-tuning and inference. We study two long-horizon manipulation domains, workspace clearing and surgical-instrument handling, which require ordered execution, visually grounded decisions, and conditional branching. We evaluate correct-object and destination selection, subtask completion, task progress, step-order consistency, complete-task success, and procedural or execution mistakes. This work positions structured symbolic reasoning and demonstration-derived visual guidance as complementary mechanisms for reliable long-horizon VLA manipulation.
\end{abstract}

\section{Introduction}

Vision-language-action (VLA) models transfer semantic knowledge from vision-language backbones into continuous robot control. OpenVLA and $\pi_{0.5}$ demonstrate broad manipulation capabilities \cite{kim2024openvla,intelligence2025pi05}, but long procedures still require persistent progress, valid dependencies, conditional branches, and reliable object and destination grounding. Collaborative progress may also depend on a human or another embodied agent.

\begin{figure}[t]
    \centering
    \includegraphics[width=\linewidth]{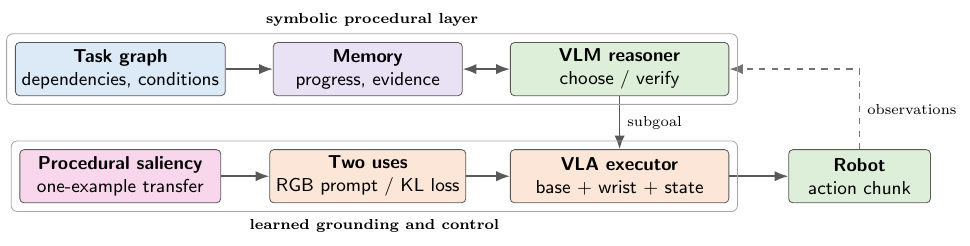}
    \caption{Neuro-symbolic procedural pipeline. The graph and event memory constrain subgoal selection, while continuous fixed-base and wrist-view perception verifies state changes caused by the robot, a human collaborator, or another agent. Sparse saliency guides the VLA through an RGB prompt or attention regularization.}
    \label{fig:framework}
\end{figure}

We address these requirements with a task graph and event memory for high-level decisions, continuous fixed-base and wrist-camera perception for transition verification, and a fine-tuned VLA for robot control. Sparse procedural saliency clarifies evidence in the stable global view. The pipeline follows an \emph{observe--ground--remember--choose--execute--verify} loop (Fig.~\ref{fig:framework}).

The approach connects neuro-symbolic and collaborative task structure
\cite{zhu2021hierarchical,paxton2017costar}, hierarchical planning and
feedback \cite{ahn2022saycan,huang2022inner,yang2025lohovla,
liu2026lohomanip}, and visual prompting and gaze-guided learning
\cite{zheng2025tracevla,wang2026vpvla,saran2021gaze,pani2026gazevla}. We contribute a graph-and-memory architecture with multiple executor roles, a shared saliency interface with explicit and internalized realizations, and initial evidence that both resolve the observed spatial-selection failures.

\section{Neuro-Symbolic Procedural Pipeline}

\subsection{Task Structure, Memory, and Collaboration}

Let $\mathcal{G}=(\mathcal{V},\mathcal{E})$ be a directed task graph. A node
\[
v_i=(\ell_i,\rho_i,\mathcal{P}^{\mathrm{pre}}_i,
\mathcal{P}^{\mathrm{succ}}_i,\mathcal{P}^{\mathrm{fail}}_i,z_i)
\]
contains an instruction $\ell_i$, executor role $\rho_i$, visually checkable conditions, and status $z_i$. Dependencies and preconditions determine readiness; $\rho_i$ may identify the robot, a human, or another embodied agent.

An append-only memory $\mathcal{M}_t=[e_1,\ldots,e_t]$ records decisions, observed transitions, and visual evidence. While acting or waiting, the system observes the fixed global view $I_t^b$ and moving wrist view $I_t^w$. A verified collaborator action becomes an external event that updates memory and can enable a robot node. For example, after the robot exposes a marked board, a human sprays it; detecting this condition releases the subsequent wiping action.

For a robot node, the dispatched subgoal conditions the low-level policy:
\[
\hat A_t=\pi_\theta(I_t^b,I_t^w,q_t,\ell_i).
\]
The graph determines \emph{what is valid next}; the VLA determines \emph{how to execute it}; perception links the two.

\subsection{Sparse Procedural Saliency}

We annotate one representative base-camera trajectory per guided subgoal with a time-varying point $g_t$. This \emph{pseudo-gaze} denotes procedural relevance rather than measured gaze. DINOv2 features \cite{oquab2023dinov2} provide temporal descriptors
and dense spatial correspondence for transfer to repeated demonstrations,
followed by pyramidal Lucas--Kanade optical-flow stabilization
\cite{lucas1981iterative}.

The signal supports two policy interfaces. \emph{Prompt-as-input} renders a hollow magenta ring $\mathcal{R}(I_t^b,g_t)$ in the RGB observation. \emph{Prompt-as-supervision} converts $g_t$ into a token-level target distribution $G_t$ and aligns action-conditioned attention $S_t$ during fine-tuning:
\[
\hat A_t^{\mathrm{vis}}
=\pi_{\theta_{\mathrm{vis}}}(\mathcal{R}(I_t^b,g_t),I_t^w,q_t,\ell_i),
\qquad
\mathcal{L}_{\mathrm{reg}}
=\mathcal{L}_{\mathrm{act}}+\lambda D_{\mathrm{KL}}(G_t\Vert S_t).
\]
The regularized policy receives clean images at inference. In the present study, both interfaces clarify spatial relationships represented in the demonstrations; they do not yet establish arbitrary redirection to unseen coordinates.

\begin{figure}[t]
    \centering
    \begin{minipage}[t]{0.485\textwidth}
        \centering
        \includegraphics[width=\linewidth]{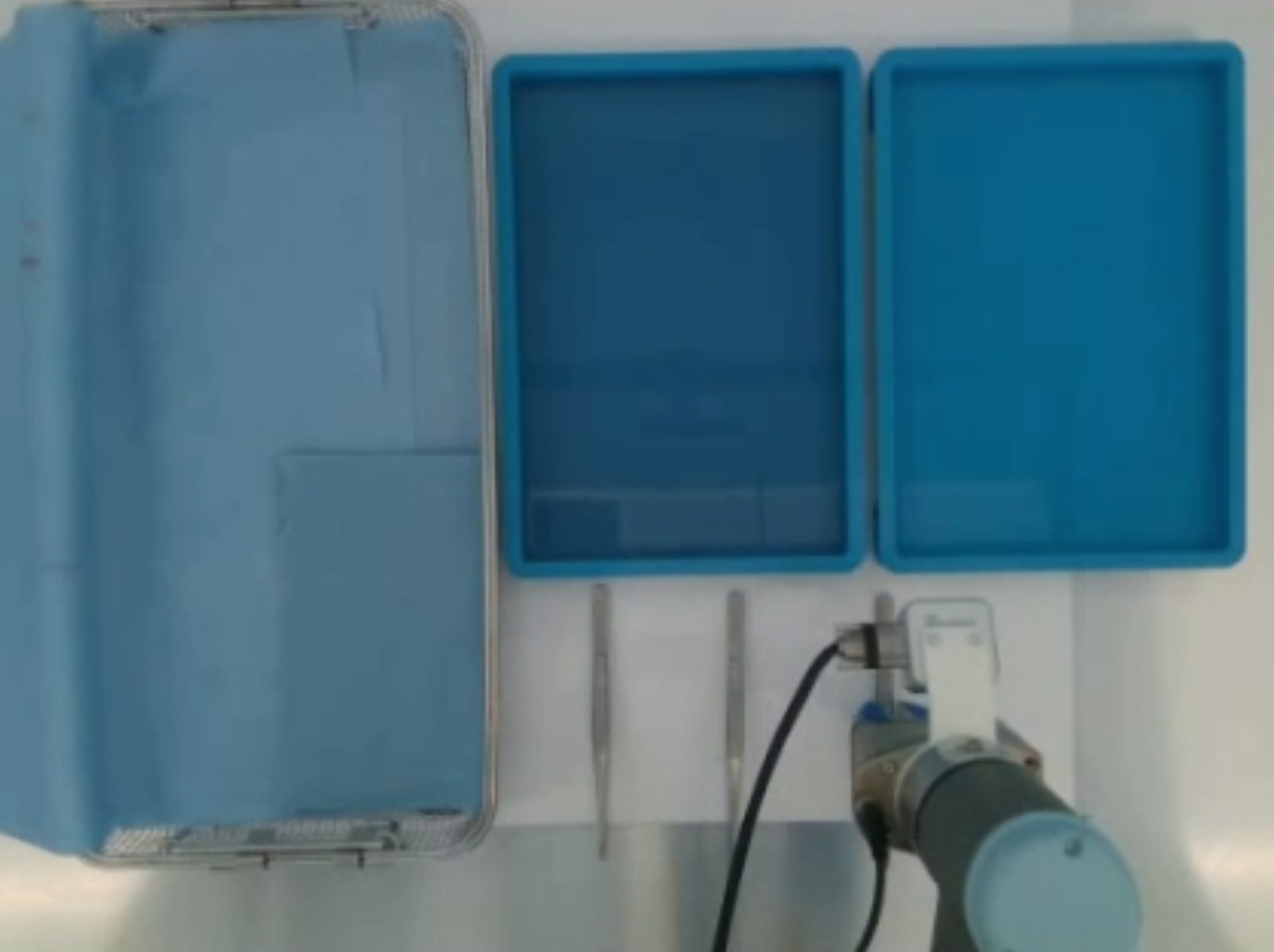}\\[-1mm]
        {\small (a) Raw base-camera observation}
    \end{minipage}\hfill
    \begin{minipage}[t]{0.485\textwidth}
        \centering
        \includegraphics[width=\linewidth]{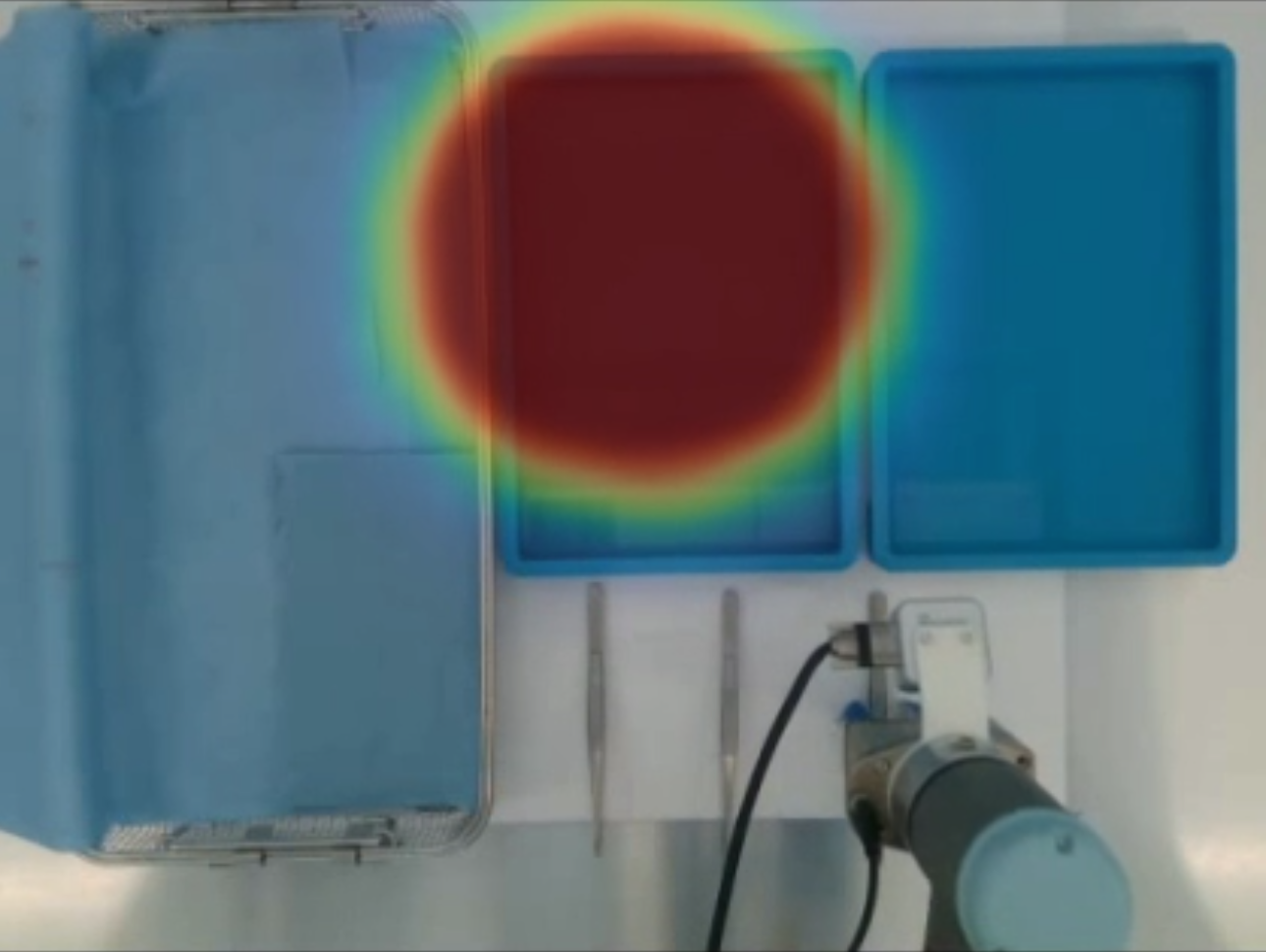}\\[-1mm]
        {\small (b) Attention heatmap overlay}
    \end{minipage}
    \caption{Representative action-conditioned attention visualization. Left: the unmodified observation. Right: the same frame with the attention-regularized policy's heatmap overlaid. The concentration around the task-relevant region illustrates alignment with procedural saliency; it serves as a qualitative diagnostic.}
    \label{fig:attention}
\end{figure}

\section{Initial Grounding Study}

We fine-tune $\pi_{0.5}$ \cite{intelligence2025pi05} from 231 teleoperated episodes of a seven-step instrument-handling procedure observed by fixed-base and wrist cameras. Five subgoals are scored under three conditions: a prompt-trained checkpoint without its expected overlay (\texttt{prompt-finetuned}), the same checkpoint with a \texttt{visual-prompt}, and a clean-input \texttt{regularized} checkpoint. Five trials per subgoal and condition yield 75 trials.

\begin{table}[t]
\centering
\caption{Initial real-robot results. Container scores aggregate the two spatially ambiguous routing subgoals. ``Highlighted'' denotes the generic instruction whose destination is supplied by the RGB cue.}
\label{tab:results}
\setlength{\tabcolsep}{3.3pt}
\begin{tabular}{lcccc}
\toprule
Condition & RGB cue & Destination text & All & Container \\
\midrule
Prompt-finetuned & none & highlighted & 20/25 & 5/10 \\
 + Visual-prompting  & ring & highlighted & 25/25 & 10/10 \\
 Regularization    & none & left/right & 25/25 & 10/10 \\
\bottomrule
\end{tabular}
\end{table}

Both guided variants complete every trial (Table~\ref{tab:results}). Removing the expected cue reduces ambiguous container routing from 10/10 to 5/10; all five failures are wrong-destination placements. The methods tie behaviorally at the current ceiling but are mechanistically distinct: prompting supplies saliency in RGB, whereas regularization yields the target-aligned pattern in Fig.~\ref{fig:attention} with clean-image inference. Prompting improved behavior without an equally pronounced aggregate attention shift, so behavioral utility and visible alignment capture different properties.

\section{Discussion and Outlook}

Explicit prompting retains a visible spatial cue at deployment; regularization internalizes the supervision and removes the overlay. Base-only guidance is sufficient in the present multi-view policy, plausibly because the global view resolves spatial ambiguity while the wrist view supports close-range control. It does not establish geometric saliency transfer into wrist coordinates.

The current score is saturated, making generalization the next question. Reversing source--destination geometry, independently moving objects and containers, and testing held-out scenes will distinguish reusable spatial grounding from strengthened familiar trajectories. Camera perturbations can then quantify view-specific contributions.

The broader pipeline remains to be evaluated end to end. Future work will measure graph-valid ordering, recognition of human-completed dependencies, transition verification, recovery, and full-task success. Human egocentric gaze can replace pseudo-gaze through cross-view transfer without changing either policy interface. Symbolic structure can thereby coordinate \emph{what} happens next, while sparse visual guidance improves \emph{how} the current subgoal is grounded.

\section*{Acknowledgements}
 This work was supported by the Fraunhofer InternaI Programs under Grant No. SME 40-12767

\bibliographystyle{splncs04}
\bibliography{main}

@inproceedings{kim2024openvla,
  title={{OpenVLA}: An Open-Source Vision-Language-Action Model},
  author={Kim, Moo Jin and Pertsch, Karl and Karamcheti, Siddharth and Xiao, Ted and Balakrishna, Ashwin and Nair, Suraj and Rafailov, Rafael and Foster, Ethan P. and Sanketi, Pannag R. and Vuong, Quan and Kollar, Thomas and Burchfiel, Benjamin and Tedrake, Russ and Sadigh, Dorsa and Levine, Sergey and Liang, Percy and Finn, Chelsea},
  booktitle={Proceedings of the 8th Conference on Robot Learning},
  series={Proceedings of Machine Learning Research},
  volume={270},
  pages={2679--2713},
  publisher={PMLR},
  year={2025},
  url={https://proceedings.mlr.press/v270/kim25c.html}
}

@inproceedings{intelligence2025pi05,
  title={$\pi_{0.5}$: A Vision-Language-Action Model with Open-World Generalization},
  author={Black, Kevin and Brown, Noah and Darpinian, James and Dhabalia, Karan and Driess, Danny and Esmail, Adnan and Equi, Michael Robert and Finn, Chelsea and Fusai, Niccolo and Galliker, Manuel Y. and Ghosh, Dibya and Groom, Lachy and Hausman, Karol and Ichter, Brian and Jakubczak, Szymon and Jones, Tim and Ke, Liyiming and LeBlanc, Devin and Levine, Sergey and Li-Bell, Adrian and Mothukuri, Mohith and Nair, Suraj and Pertsch, Karl and Ren, Allen Z. and Shi, Lucy Xiaoyang and Smith, Laura and Springenberg, Jost Tobias and Stachowicz, Kyle and Tanner, James and Vuong, Quan and Walke, Homer and Walling, Anna and Wang, Haohuan and Yu, Lili and Zhilinsky, Ury},
  booktitle={Proceedings of the 9th Conference on Robot Learning},
  series={Proceedings of Machine Learning Research},
  volume={305},
  pages={17--40},
  publisher={PMLR},
  year={2025},
  url={https://proceedings.mlr.press/v305/black25a.html}
}

@inproceedings{ahn2022saycan,
  title={Do As I Can, Not As I Say: Grounding Language in Robotic Affordances},
  author={Ichter, Brian and Brohan, Anthony and Chebotar, Yevgen and Finn, Chelsea and Hausman, Karol and Herzog, Alexander and Ho, Daniel and Ibarz, Julian and Irpan, Alex and Jang, Eric and Julian, Ryan and Kalashnikov, Dmitry and Levine, Sergey and Lu, Yao and Parada, Carolina and Rao, Kanishka and Sermanet, Pierre and Toshev, Alexander T. and Vanhoucke, Vincent and Xia, Fei and Xiao, Ted and Xu, Peng and Yan, Mengyuan and Brown, Noah and Ahn, Michael and Cortes, Omar and Sievers, Nicolas and Tan, Clayton and Xu, Sichun and Reyes, Diego and Rettinghouse, Jarek and Quiambao, Jornell and Pastor, Peter and Luu, Linda and Lee, Kuang-Huei and Kuang, Yuheng and Jesmonth, Sally and Joshi, Nikhil J. and Jeffrey, Kyle and Ruano, Rosario Jauregui and Hsu, Jasmine and Gopalakrishnan, Keerthana and David, Byron and Zeng, Andy and Fu, Chuyuan Kelly},
  booktitle={Proceedings of the 6th Conference on Robot Learning},
  series={Proceedings of Machine Learning Research},
  volume={205},
  pages={287--318},
  publisher={PMLR},
  year={2023},
  url={https://proceedings.mlr.press/v205/ichter23a.html}
}

@inproceedings{huang2022inner,
  title={Inner Monologue: Embodied Reasoning through Planning with Language Models},
  author={Huang, Wenlong and Xia, Fei and Xiao, Ted and Chan, Harris and Liang, Jacky and Florence, Pete and Zeng, Andy and Tompson, Jonathan and Mordatch, Igor and Chebotar, Yevgen and Sermanet, Pierre and Jackson, Tomas and Brown, Noah and Luu, Linda and Levine, Sergey and Hausman, Karol and Ichter, Brian},
  booktitle={Proceedings of the 6th Conference on Robot Learning},
  series={Proceedings of Machine Learning Research},
  volume={205},
  pages={1769--1782},
  publisher={PMLR},
  year={2023},
  url={https://proceedings.mlr.press/v205/huang23c.html}
}

@article{yang2025lohovla,
  title={{LoHoVLA}: A Unified Vision-Language-Action Model for Long-Horizon Embodied Tasks},
  author={Yang, Yi and Sun, Jiaxuan and Kou, Siqi and Wang, Yihan and Deng, Zhijie},
  journal={arXiv preprint arXiv:2506.00411},
  year={2025},
  url={https://arxiv.org/abs/2506.00411}
}

@article{liu2026lohomanip,
  title={Long-Horizon Manipulation via Trace-Conditioned {VLA} Planning},
  author={Liu, Isabella and Cheng, An-Chieh and Yan, Rui and Chen, Geng and Qiu, Ri-Zhao and Zou, Xueyan and Yi, Sha and Yin, Hongxu and Wang, Xiaolong and Liu, Sifei},
  journal={arXiv preprint arXiv:2604.21924},
  year={2026},
  url={https://arxiv.org/abs/2604.21924}
}

@inproceedings{zheng2025tracevla,
  title={{TraceVLA}: Visual Trace Prompting Enhances Spatial-Temporal Awareness for Generalist Robotic Policies},
  author={Zheng, Ruijie and Liang, Yongyuan and Huang, Shuaiyi and Gao, Jianfeng and Daum{\'e} III, Hal and Kolobov, Andrey and Huang, Furong and Yang, Jianwei},
  booktitle={International Conference on Learning Representations},
  year={2025},
  url={https://openreview.net/forum?id=b1CVu9l5GO}
}

@article{wang2026vpvla,
  title={{VP-VLA}: Visual Prompting as an Interface for Vision-Language-Action Models},
  author={Wang, Zixuan and Chen, Yuxin and Liu, Yuqi and Ye, Jinhui and Chen, Pengguang and Lu, Changsheng and Liu, Shu and Jia, Jiaya},
  journal={arXiv preprint arXiv:2603.22003},
  year={2026},
  url={https://arxiv.org/abs/2603.22003}
}

@inproceedings{saran2021gaze,
  title={Efficiently Guiding Imitation Learning Agents with Human Gaze},
  author={Saran, Akanksha and Zhang, Ruohan and Short, Elaine Schaertl and Niekum, Scott},
  booktitle={Proceedings of the 20th International Conference on Autonomous Agents and Multiagent Systems},
  pages={1109--1117},
  publisher={IFAAMAS},
  year={2021},
  url={https://www.ifaamas.org/Proceedings/aamas2021/pdfs/p1109.pdf}
}

@inproceedings{pani2026gazevla,
  title={Gaze-Regularized Vision-Language-Action Models for Robotic Manipulation},
  author={Pani, Anupam and Yang, Yanchao},
  booktitle={Proceedings of the IEEE/CVF Conference on Computer Vision and Pattern Recognition Workshops},
  pages={11332--11342},
  year={2026},
  url={https://openaccess.thecvf.com/content/CVPR2026W/GRAIL-V/html/Pani_Gaze-Regularized_Vision-Language-Action_Models_for_Robotic_Manipulation_CVPRW_2026_paper.html}
}

@article{oquab2023dinov2,
  title={{DINOv2}: Learning Robust Visual Features without Supervision},
  author={Oquab, Maxime and Darcet, Timoth{\'e}e and Moutakanni, Th{\'e}o and Vo, Huy V. and Szafraniec, Marc and Khalidov, Vasil and Fernandez, Pierre and Haziza, Daniel and Massa, Francisco and El-Nouby, Alaaeldin and Assran, Mahmoud and Ballas, Nicolas and Galuba, Wojciech and Howes, Russell and Huang, Po-Yao and Li, Shang-Wen and Misra, Ishan and Rabbat, Michael and Sharma, Vasu and Synnaeve, Gabriel and Xu, Hu and J{\'e}gou, Herv{\'e} and Mairal, Julien and Labatut, Patrick and Joulin, Armand and Bojanowski, Piotr},
  journal={Transactions on Machine Learning Research},
  year={2024},
  url={https://openreview.net/forum?id=a68SUt6zFt}
}

@inproceedings{lucas1981iterative,
  title={An Iterative Image Registration Technique with an Application to Stereo Vision},
  author={Lucas, Bruce D. and Kanade, Takeo},
  booktitle={Proceedings of the 7th International Joint Conference on Artificial Intelligence},
  pages={674--679},
  year={1981}
}

@inproceedings{zhu2021hierarchical,
  title={Hierarchical Planning for Long-Horizon Manipulation with Geometric and Symbolic Scene Graphs},
  author={Zhu, Yifeng and Tremblay, Jonathan and Birchfield, Stan and Zhu, Yuke},
  booktitle={2021 IEEE International Conference on Robotics and Automation},
  pages={6541--6548},
  publisher={IEEE},
  year={2021},
  doi={10.1109/ICRA48506.2021.9561548}
}

@inproceedings{paxton2017costar,
  title={{CoSTAR}: Instructing Collaborative Robots with Behavior Trees and Vision},
  author={Paxton, Chris and Hundt, Andrew and Jonathan, Felix and Guerin, Kelleher and Hager, Gregory D.},
  booktitle={2017 IEEE International Conference on Robotics and Automation},
  pages={564--571},
  publisher={IEEE},
  year={2017},
  doi={10.1109/ICRA.2017.7989070}
}
\end{document}